\documentclass[10pt,twocolumn,letterpaper]{article}

\usepackage{cvpr}
\usepackage{times}
\usepackage{epsfig}
\usepackage{graphicx}
\usepackage{amsmath}
\usepackage{amssymb}
\usepackage{subcaption}
\usepackage{multirow}

\usepackage[pagebackref=true,breaklinks=true,letterpaper=true,colorlinks,bookmarks=false]{hyperref}

\cvprfinalcopy 

\def\cvprPaperID{} 

\ifcvprfinal\fi
\begin{document}

\title{A Baseline for the Commands For Autonomous Vehicles Challenge.}

\author{Simon Vandenhende \\
KU Leuven - ESAT/PSI\\
{\tt\small simon.vandenhende@kuleuven.be}
\and
Thierry Deruyttere\\
KU Leuven - LIIR\\
{\tt\small thierry.deruyttere@kuleuven.be}
\and
Dusan Grujicic\\
KU Leuven - ESAT/PSI\\
{\tt\small dusan.grujicic@kuleuven.be}
}

\maketitle

\begin{abstract}
The \textit{Commands For Autonomous Vehicles} (C4AV) challenge requires participants to solve an object referral task in a real-world setting. More specifically, we consider a scenario where a passenger can pass free-form natural language commands to a self-driving car. This problem is particularly challenging, as the language is much less constrained compared to existing benchmarks, and object references are often implicit. The challenge is based on the recent \texttt{Talk2Car} dataset. This document provides a technical overview of a model that we released to help participants get started in the competition. The code can be found at  \url{https://github.com/talk2car/Talk2Car}.
\end{abstract}

\section{Introduction}
\begin{figure*}
\begin{subfigure}[t]{.32\textwidth}
  \centering
  \includegraphics[width=\linewidth]{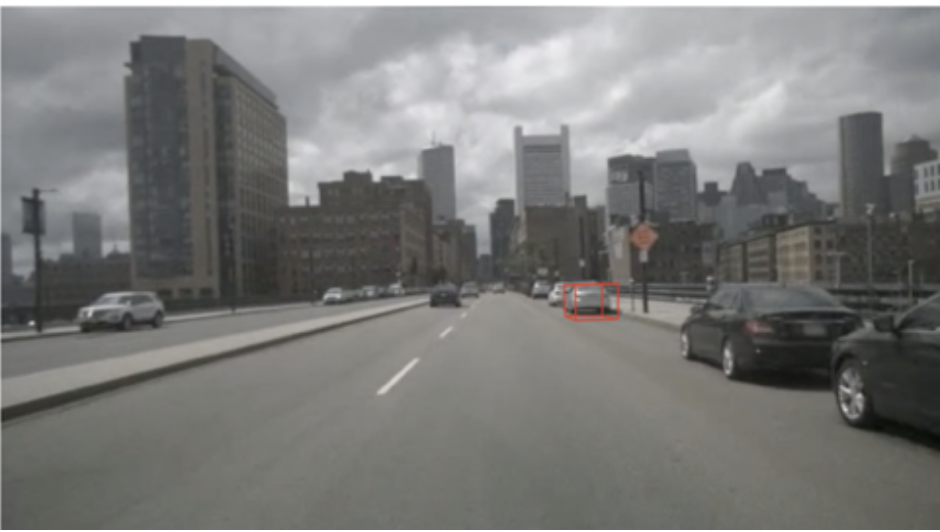}
  \caption{You can park up ahead behind \textbf{the silver car}, next to that lamppost with the orange sign on it.}
\end{subfigure}%
\hfill
\begin{subfigure}[t]{.32\textwidth}
  \centering
  \includegraphics[width=\linewidth]{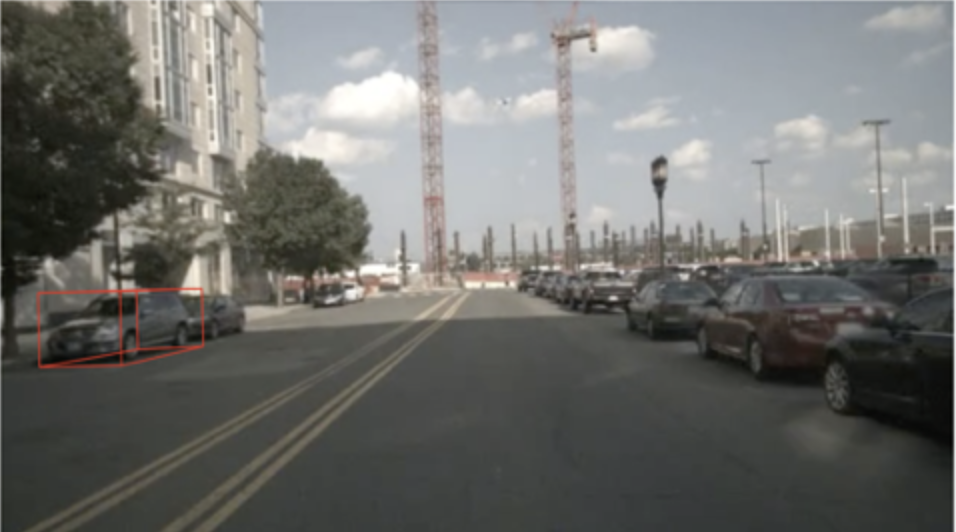}
  \caption{Turn around and park in front of \textbf{that vehicle in the shade}.}
\end{subfigure}
\hfill
\begin{subfigure}[t]{.32\textwidth}
  \centering
  \includegraphics[width=\linewidth]{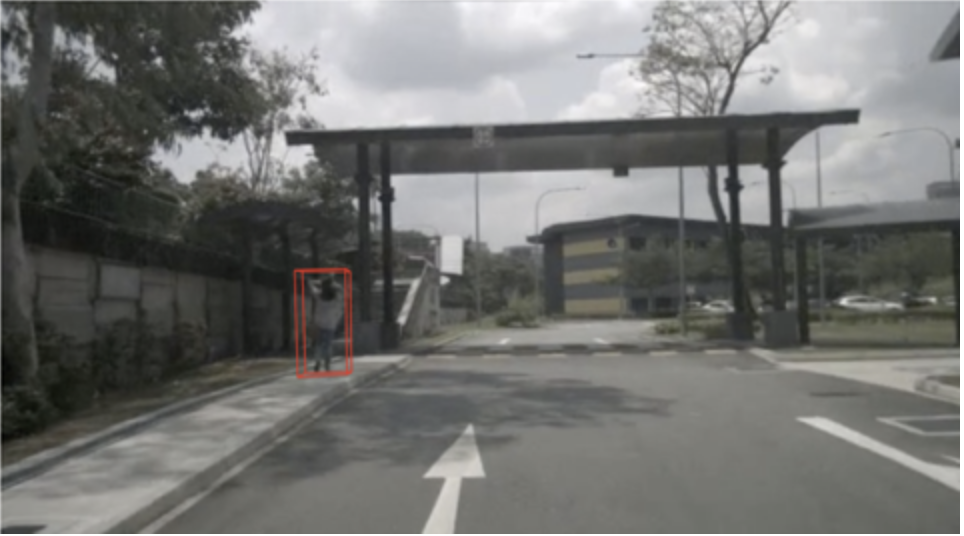}
  \caption{There's \textbf{my mum}, on the left! The one walking closest to us. Park near \textbf{her}, she might want a lift.}
\end{subfigure}
\caption{Examples from the \texttt{Talk2Car} dataset. Each command describes a change of direction for the vehicle, relevant to a referred object found in the scene (here indicated by the red 3D-bounding box). The C4AV challenge requires to predict the correct bounding box from the image and the command.}
\label{fig: examples}
\end{figure*}

The \textit{Commands For Autonomous Vehicles} (C4AV) challenge is build on top of the \textit{Talk2Car} dataset~\cite{talk2car}. Given a command $C$ and an image $I$, we are required to predict the coordinates of a bounding box drawn around the object the command is referring to. Some examples can be found in Figure~\ref{fig: examples}. The model is evaluated based on its AP50 score. More details about the benchmark and the dataset are provided in the original works~\cite{talk2car,nuscenes}.

The remainder of this text discusses the implementation of a  baseline model to help participants get started in the competition. The model was implemented in PyTorch~\cite{pytorch}.

\begin{figure*}
    \centering
    \includegraphics[width=\linewidth]{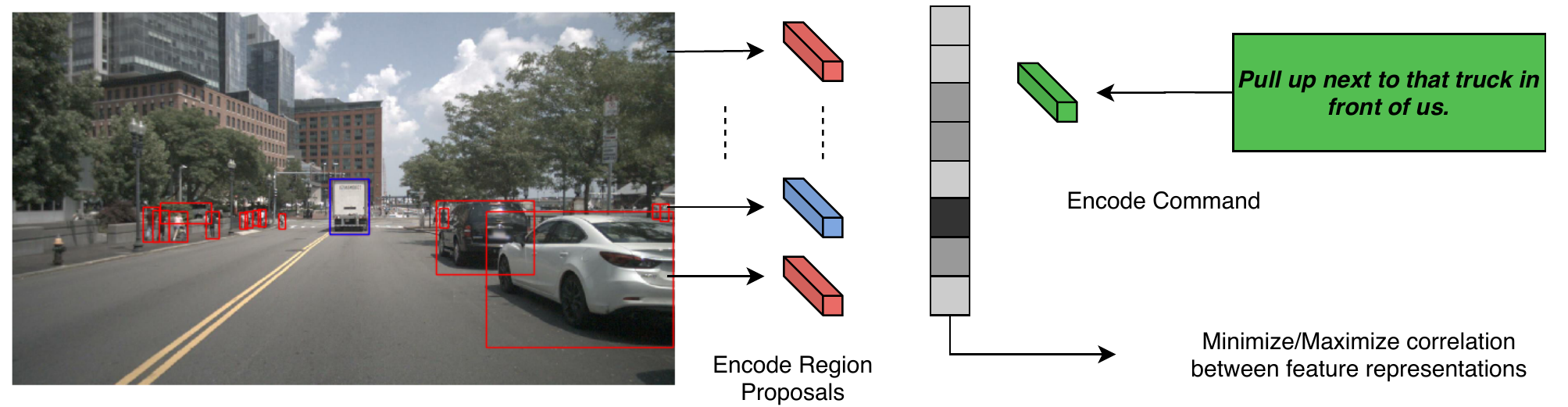}
    \caption{Overview of the method. First, we extract object region proposals. Second, every region proposal is encoded using an off-the-shelf image encoder. In parallel, we encode the command using a recurrent neural network. Finally, we measure the correlation between the feature representations of the region proposals and the command. The training minimizes or maximizes the feature correlations between the regions and the commands depending on whether the object of interest is included in the bounding box ($IoU > 0.5$) or not.}
    \label{fig: method}
\end{figure*}

\section{Model}
\subsection{Overview}
Figure~\ref{fig: method} shows an overview of the method. The procedure is similar to the work from Karpathy et al.~\cite{karpathy2014deep}. First, an object detector is used to extract region proposals from the input image. Second, we match the region proposals with the command. In particular, we compute the feature correlations between an encoding of the region proposals and the command. The feature correlations are considered as a score for how well the bounding box fits the command. The model is trained by mapping the feature correlations to zero or one, depending on whether the proposed region has an IoU overlap of at least 0.5 with the ground-truth bounding box. 

\subsection{Implementation Details}
\subsubsection{Region Proposals}
We obtain 64 region proposals for every image by finetuning the CenterNet~\cite{zhou2019objects} model on the object detection task. Notice that we only use images from the \texttt{Talk2Car} dataset. The annotations are provided by nuScenes~\cite{nuscenes}. We deliver the region proposals for every image as part of the code package. Duplicate regions are removed through non-maximum suppression. 

\subsubsection{Method}
For every image, we select the 16 regions with the highest confidence score. The region proposals are divided into a set of positive ($IoU \geq 0.5$) and negative boxes ($IoU < 0.5)$. A ResNet-18 model that was pretrained on ImageNet is used to encode the regions. The command is encoded using a bi-directional GRU. The feature correlations are measured after L2 normalization and linearly remapped to lie between 0 and 1. The loss for a sample, which is an image with a command and bounding boxes, is calculated with a binary cross-entropy as follows
\begin{equation}
  -\frac{1}{|\mathcal{P}|}  \sum_{x \in \mathcal{P}} x \log x - \frac{1}{|\mathcal{N}|} \sum_{x \in \mathcal{N}} (1-x) \log(1-x)
\end{equation}
where $\mathcal{P}$ contains the feature correlations for positive boxes, and $\mathcal{N}$ contains the feature correlations for negative boxes. As most region proposals do not overlap with the ground truth bounding box, the number of negative boxes greatly exceeds the number of positive boxes. Therefore, we average the loss of the positive and negative boxes separately, in order to avoid the network from getting biased towards negative predictions.

We train the model for 10 epochs using SGD with nesterov momentum $0.9$. The initial learning is set to 0.01 and decayed by a factor 10 every 4 epochs. We use batches of size 18 and a weight decay term 1e-4. The model can be trained in two hours on a single Nvidia 1080ti GPu. Our best model obtains $43.5\%$ AP50 on the validation set.

{\small
\bibliographystyle{ieee_fullname}
\bibliography{bibliography}
}

\end{document}